# Knowledge Based Wind Turbine Gearbox Fault Detection Using Sparse Filtering and Graph Neural Networks

Jinsong Wang and Kenneth A. Loparo

Department of Electrical, Computer, and Systems Engineering, Case Western Reserve University, Cleveland, Ohio 44106, USA

## Abstract

Reliable gearbox condition monitoring requires methods that can process high-frequency vibration measurements while retaining information about sensor locations and drivetrain structure. We investigate a two-stage framework that combines sparse filtering with graph neural networks for wind turbine gearbox fault detection. Sparse filtering learns compact features from vibration segments, and a knowledge-based graph associates these features with gearbox components, sensors, and health-state entities. A basic graph neural network and a gated graph neural network produce a single health prediction, whereas a gated graph sequence neural network generates structured sequences that associate a predicted state with component and sensor information. Experiments use eight accelerometer channels from the National Renewable Energy Laboratory gearbox condition-monitoring benchmark. With 100 learned features, the sparse-filtering gated graph neural network achieves a reported classification accuracy of 93.74%, compared with 89.28% for its manually engineered feature counterpart. The sequence model achieves reported accuracies of 90.73% and 90.36% in the one- and two-fault-signal cases, respectively. These results support the use of learned vibration representations within a structured sensor graph. The evidence is restricted to the evaluated benchmark and operating condition; independent component localization, transfer across turbines, and deployment latency require further validation.

## 1 Introduction

### 1.1 Gearbox monitoring and the diagnostic problem

Gearbox faults can interrupt wind turbine operation and require substantial maintenance effort. Condition monitoring therefore seeks to identify changes in drivetrain behavior early enough to support inspection and maintenance planning. Vibration measurements are particularly useful because gear meshing, bearing interactions, and mechanical defects produce dynamic responses that can be observed by accelerometers. Interpreting those measurements remains difficult: signals depend on sensor position, transmission paths, shaft speed, load, and the interaction of multiple gearbox elements. A diagnostic model must distinguish changes associated with damage from changes associated with normal operation.

Monitoring also involves several different inference tasks. Fault detection identifies whether a gearbox departs from a healthy condition. Diagnosis additionally identifies the affected component or fault mechanism. Prognosis estimates future degradation or remaining life. These tasks require different labels and validation procedures. In this study, the primary quantitative task is supervised discrimination between healthy and damaged gearbox measurements. Structured sequence outputs additionally associate predictions with component and sensor entities. Such associations can make an output easier to interpret, but their diagnostic meaning depends on the availability of independent component-level labels.

An instrumented gearbox is naturally a multivariate system. Accelerometers mounted near different shafts, bearings, and gear stages provide related observations of the same drivetrain. Processing each channel independently can overlook these relationships. Concatenating channels allows a conventional classifier to use multiple signals, but does not explicitly describe where the sensors are mounted or how the corresponding components are connected. A graph provides a way to encode those relationships alongside the measured data. The resulting model can combine signal features with a representation of the monitored equipment.

### 1.2 Background and related work

Earlier intelligent diagnosis methods commonly separate signal processing, feature extraction, and classification. Time-domain statistics summarize amplitude distributions and impulsiveness, while frequency and time-frequency representations describe periodic components and nonstationary behavior. Samanta combined neural networks or support vector machines with genetic feature selection for gear fault classification [1]. Amar et al. used vibration spectrum imaging for bearing fault classification [2]. These studies illustrate how diagnostic performance depends on both the representation and the classifier. Their results concern their own datasets and should not be interpreted as direct performance benchmarks for the present gearbox experiment.

Wind turbine monitoring also uses lower-rate supervisory control and data acquisition measurements. Pandit and Infield developed Gaussian-process models for SCADA-based anomaly detection [3], while Bangalore and Tjernberg investigated neural-network detection of gearbox bearing faults [4]. Other studies have used tree ensembles, denoising autoencoders, and combinations of signal decomposition methods [5–7]. These approaches address complementary aspects of the monitoring problem. SCADA-based models describe relationships among operating variables, whereas high-frequency vibration analysis resolves mechanical responses at a much finer sampling scale. The choice of data modality therefore affects both the detectable phenomena and the computational requirements.

Feature learning reduces reliance on a manually selected set of statistics. Sparse filtering learns a representation by optimizing the sparsity of normalized feature activations, rather than reconstructing the input signal [8]. Lei et al. applied unsupervised feature learning to mechanical fault diagnosis [9]. This provides the motivation for learning a compact representation before graph processing. However, an unsupervised sparsity objective does not by itself guarantee noise suppression or preservation of fault information. Its diagnostic value must be assessed through the downstream task, with preprocessing fitted independently of the test set.

Graph neural networks extend representation learning to relational data. The original GNN formulation updates node states using neighboring nodes and edge information [10]. Gated graph neural networks introduce recurrent gates and a fixed number of propagation steps, and gated graph sequence neural networks extend the output to sequences [11]. These models are relevant when a monitoring system contains heterogeneous entities and typed relationships. Ontologies provide a complementary representation of domain concepts; OntoWind, for example, organizes wind-energy terminology and semantic relationships [12]. An ontology can guide graph construction, while a GNN learns how node attributes and connections relate to the prediction target.

Recent wind turbine studies have substantially expanded this research direction. Tan et al. investigated graph attention networks for gearbox vibration diagnosis [13]. Guo et al. used graph contrastive learning for semi-supervised diagnosis with limited labeled SCADA data [14]. A 2025 study developed a spatio-temporal graph convolutional network to model dependencies in SCADA measurements [15]. These approaches show that graph construction, temporal representation, and label availability are distinct design choices. Their use of learned or similarity-based relationships also differs from a graph whose structure is specified through equipment knowledge.

Research published in 2026 further emphasizes the integration of domain knowledge and temporal learning. Kim and Lee combined a maintenance knowledge graph with time-series relational graph convolution for gearbox predictive maintenance using simulation-generated data [16]. Another study combined FFT features, GraphSAGE, graph attention, and genetic optimization on NREL gearbox measurements [17]. Fu et al. investigated spatio-temporal graph modeling under time-varying operating conditions [18]. These developments place the present framework within a growing literature on graph-based wind turbine monitoring and establish relevant comparison directions, although numerical rankings require matched datasets, channel selections, data partitions, and evaluation metrics.

Generalization remains a separate challenge from accuracy within one benchmark. A model trained at one speed and load can encounter different signal distributions during operation. Recent work has examined adversarial learning for speed–load variation [19] and uncertainty-aware detection of out-of-distribution gearbox measurements [20]. These issues are particularly relevant when healthy and damaged data come from different physical gearboxes. A classifier may exploit specimen-specific characteristics as well as damage-related features, so independent validation is needed before claiming transfer to previously unseen equipment.

### 1.3 Study objective and contributions

We investigate whether compact features learned from high-frequency vibration signals can improve fault detection within a knowledge-based graph of a wind turbine gearbox. The framework, termed SF-GNN, separates unsupervised feature learning from supervised graph prediction. The same equipment representation supports a single health-state output and a structured sequence output. The intended contribution is this integration and its evaluation on the selected benchmark, rather than a new sparse-filtering objective or a new graph propagation architecture.

The study contributes a graph representation linking gearbox components, sensor channels, and health states; a two-stage implementation using sparse filtering with basic, gated, and sequence graph models; and an empirical comparison of feature dimension, single-output accuracy, sequence-output accuracy, and reported running time. The experiments also reveal the limits of the available evidence. In particular, the results support gearbox condition classification under the tested setting, while the interpretation of sequence predictions as physical fault localization requires additional ground truth.

The remainder of this paper is organized as follows. Section 2 introduces the learning formulations. Section 3 describes graph construction and the proposed workflow. Section 4 presents the dataset and experimental configuration. Section 5 reports the results and discusses their implications and limitations. Section 6 concludes the paper.

## 2 Learning formulations

### 2.1 Graph representation and node prediction

Let G = (V, E, R) be a directed, typed graph, where V is the node set, E is the edge set, and R is the set of relation types. A directed edge (u, r, v) links source node u to target node v through relation r. Each node has an observed attribute vector $x_v$ and a learned hidden state $h_v$. Node attributes describe entity types and, for sensor nodes, features extracted from the associated vibration segment. We reserve y for a supervised target label, so that observed attributes and unknown health labels are not conflated.

In a generic message-passing formulation, a node receives messages from its neighbors and updates its state. A compact description is

$$m_v^{(t)} = \Sigma_{(u,r,v) \in E} M_r(h_u^{(t-1)}, x_u, x_v)$$
$$h_v^{(t)} = U(h_v^{(t-1)}, m_v^{(t)}) \quad (1)$$

Here $M_r$ is a message function and U is a state-update function. Incoming and outgoing edges can be represented separately, allowing direction to affect propagation. Equation (1) describes the model family and does not prescribe additional architectural choices beyond the experiment. In a basic recurrent GNN, transition functions share parameters across updates [10]. This formulation should be distinguished from a gated model with a fixed propagation depth.

After propagation, the state of a designated prediction node is mapped to class probabilities:

$$p(y_v = c \mid G, X) = softmax(g(h_v^{(T)}, x_v))_c \quad (2)$$

The readout g is a trainable output function and T denotes the final propagation step. For the single-output experiment, the prediction represents gearbox health. The graph includes candidate health-state entities, but the true class is a training target and must not be available as an input attribute or a sample-dependent edge at evaluation time. Otherwise, the graph would reveal the answer rather than require inference from vibration features.

## 2.2 Gated propagation and sequence prediction

A gated graph neural network uses a gated recurrent unit to update node states [11]. Its propagation can be summarized as

$$h_v^{(t)} = GRU(m_v^{(t)}, h_v^{(t-1)}), \quad t = 1, \dots, T \quad (3)$$

The initial state encodes the observed node attributes, with padding or an input mapping when necessary. The gates control how incoming information changes the existing representation. They provide a mechanism for learning recurrent updates, but do not guarantee freedom from overfitting or optimization difficulties. Propagation depth, hidden-state dimension, and parameterization of relation types remain important implementation choices.

A gated graph sequence neural network generates an ordered output by alternating graph propagation, prediction, and annotation updates. If q = ($q_1$, …, $q_L$) is the target sequence, its conditional probability is

$$p(q \mid G, X) = \Pi_{k=1}^{L} p(q_k \mid q_1, \dots, q_{k-1}, G, X) \quad (4)$$

In the present application, sequence elements describe components, sensors, relations, and a health state. The sequence index k represents a decoding step; it is not a physical time index. Consequently, sequence generation should not be described as forecasting a fault trajectory. Errors in earlier predictions can influence later annotations, which provides a possible explanation for lower sequence accuracy, although a dedicated error analysis is needed to establish that mechanism.

## 2.3 Sparse filtering

For one sensor channel, let X = [$x_1$, …, $x_N$] contain N vibration segments, with each $x_n$ containing D samples. Sparse filtering learns K features through a weight matrix W. To state the normalization axes explicitly, the standard formulation uses nonnegative activations $F_{kn}$, feature-wise normalization across training examples, and then example-wise normalization across features [8]:

$$F_{kn} = \sqrt{(w_k^T x_n)^2 + \varepsilon}$$

$$F_{kn} = \frac{F_{kn}}{\sqrt{\Sigma_n F_{kn}^2 + \varepsilon}} \quad (5)$$

$$F_{kn} = \frac{F_{kn}}{\sqrt{\Sigma_k F_{kn}^2 + \varepsilon}}$$

$$W^* = arg\ min_W\ \Sigma_n\ \Sigma_k\ F_{kn} \quad (6)$$

The small positive constant ε stabilizes the computation. These equations make explicit the smooth absolute-value activation and the two normalization operations implicit in a linear feature description. The implementation must use a consistent matrix orientation and normalization convention. At test time, the learned weights and the training-derived feature normalization are held fixed; test examples must not be used to refit the representation.

The objective favors sparse activity within each example and discourages a solution dominated by a small subset of features. Population sparsity, lifetime sparsity, and dispersal describe the desired activation distribution. They do not imply that every feature has exactly the same activation or that irrelevant vibration content is necessarily removed. The number K is the principal representation-size hyperparameter. Input segmentation, optimization settings, and numerical stabilization also affect the implementation.

## 3 Proposed framework

### 3.1 Knowledge based graph construction

The graph combines equipment structure with sensor information. Its terminology is adapted from the wind-energy ontology of Küçük and Küçük [12]. Component nodes represent the turbine, gearbox, shafts, gears, and bearings. Sensor/data nodes represent the monitored channels and their measured information. State nodes represent candidate health conditions, and meta nodes identify entity or measurement types. Figure 1 illustrates a small part of this schema and distinguishes structural knowledge from a predicted association.

The relation vocabulary contains four semantic categories: has, is-a, measures, and causes. The has relation represents a structural or monitoring association; is-a identifies an entity type; measures associates a sensor with a measurement type. The original causes relation connects observations with health states. Here it is interpreted as an association used for prediction. A sensor observes a mechanical response and does not physically cause the gearbox fault. Establishing a causal mechanism would require evidence beyond the graph label.

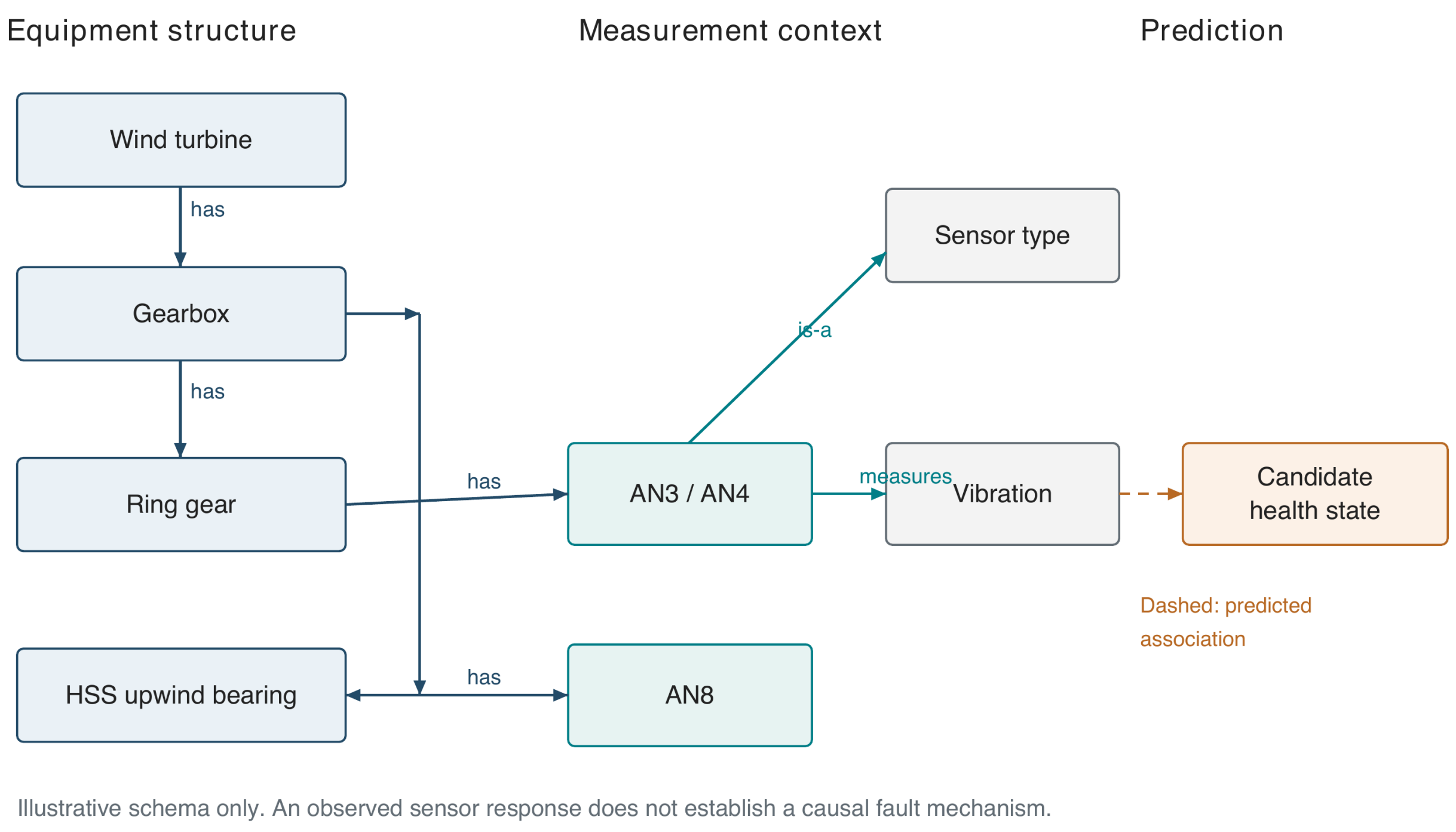

Figure 1. Illustrative subset of the knowledge-based graph. Solid arrows show equipment or measurement relationships. The dashed arrow shows a predicted association with a candidate health state; it does not establish physical causation. AN3, AN4, and AN8 are example channels. The diagram illustrates the schema rather than the complete experimental adjacency matrix.

The experimental graph is described as containing 21 terminology nodes, eight data nodes, two state nodes, and four meta nodes, for a total of 35 nodes. For each synchronized vibration segment, the topology represents the equipment configuration and the sensor-node attributes carry the corresponding signal features. The graph therefore combines fixed structural information with attributes that vary from segment to segment. Reproducing the experiment requires the full node list, edge list, edge directions, and mapping from each channel to its node.

The graph structure is a prior assumption about the monitored system. Structural connections can guide information exchange, but a useful ontology does not automatically improve classification. A matched experiment with removed, shuffled, or untyped edges is needed to measure the contribution of the chosen relationships. Similarly, static relations known from the equipment description should be distinguished from relations predicted by the sequence decoder.

### 3.2 Feature learning and graph inputs

The workflow begins by segmenting the high-frequency vibration recordings. Each sensor segment contains 4,000 samples, corresponding to 0.1 s at 40 kHz. Sparse filtering transforms the segment into a K-dimensional feature vector. For the selected setting, K = 100, so the sensor representation has 40 times fewer entries than the original segment. This is a reduction in feature dimension, not a measured 40-fold reduction in processing time.

The learned vectors are assigned to their corresponding sensor nodes. Equipment and meta nodes provide the structural context used during graph propagation. Consistent feature dimensions are required across the model inputs, but the precise padding or projection used for non-sensor nodes must be specified in the implementation. Whether sparse-filtering weights are shared across channels or fitted separately is also a reproducibility detail; these alternatives should not be treated as equivalent without verification.

Feature fitting must follow the data partition. Training recordings determine the learned representation and any data-dependent normalization. Development data support model selection, including the choice of K, and test data provide a final evaluation. Figure 2 shows the intended separation of these stages. This separation is essential even though sparse filtering does not use class labels, because access to held-out examples during feature fitting can influence the representation.

### 3.3 Single and sequential outputs

The single-output variants, SF-GNN and SF-GG-NN, classify the health condition from the attributed graph. Their difference lies in graph-state propagation, while both use sparse-filtered sensor features. The sequential variant, SF-GGS-NN, additionally produces an ordered description of selected entities and relationships. This allows a prediction to retain its association with a component and a measurement channel.

For example, a decoded sequence can connect a gearbox to a ring gear, the ring gear to channel AN3, and AN3 to a vibration measurement and a predicted damaged state. Such a sequence expresses how the model organizes the evidence. A prediction associated with AN3 does not independently prove that the ring gear is the unique damaged component: vibration can propagate through the drivetrain, and the damaged benchmark gearbox contains several defects. The sequence must therefore be evaluated against an explicit annotation scheme before it can support component-level localization claims.

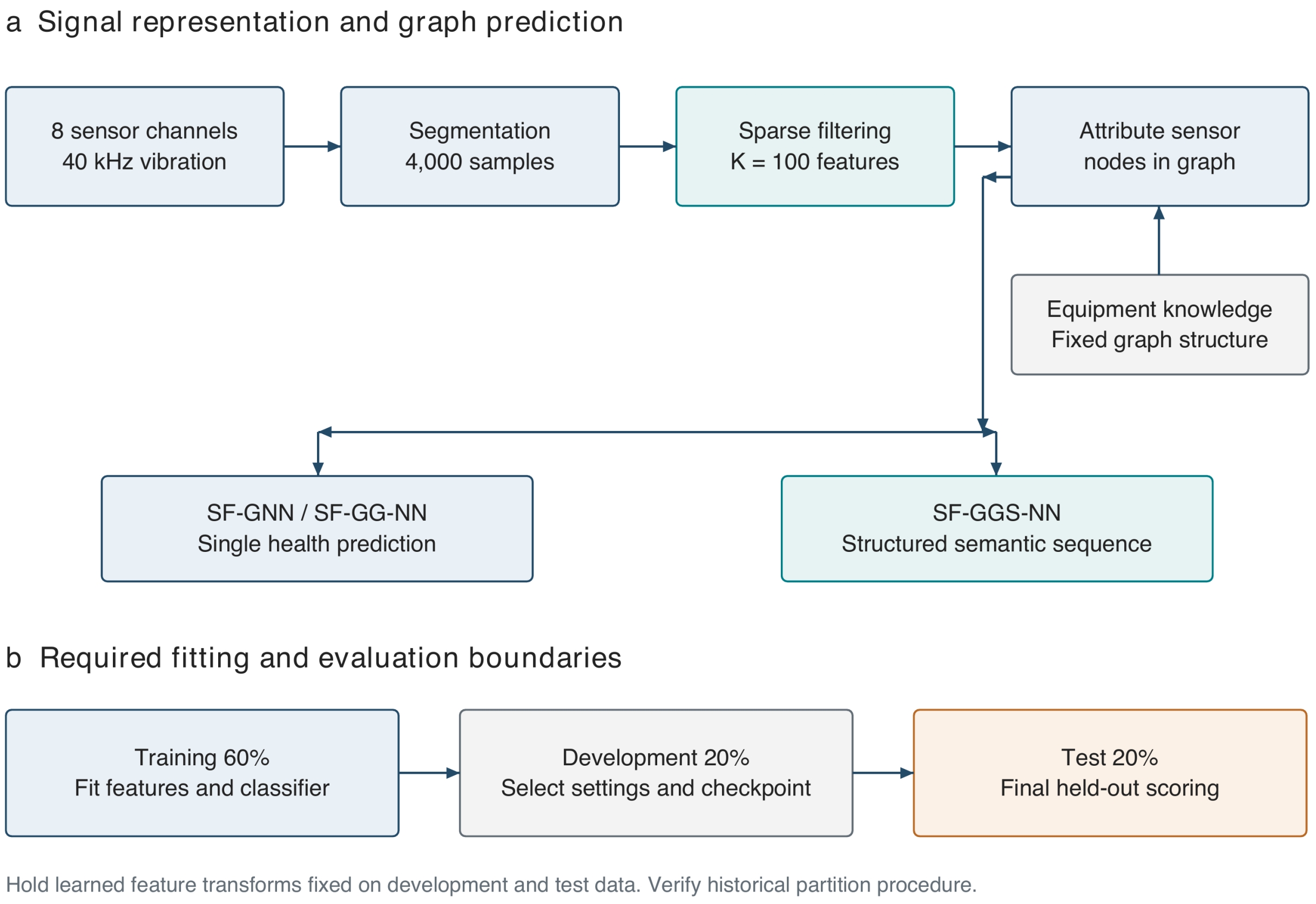

Figure 2. SF-GNN workflow and evaluation boundaries. Sparse filtering learns a compact sensor representation before supervised graph prediction. The two output branches distinguish a single health decision from a structured semantic sequence. Training-derived transformations are held fixed for development and test data. The partition boundary depicts the required evaluation procedure; the historical experiment requires verification against the implementation.

# 4 Experimental design

## 4.1 Dataset and operating condition

The experiments use the NREL wind turbine gearbox vibration condition-monitoring benchmark [21]. It contains healthy and damaged gearboxes of the same design. The damaged unit experienced oil-loss events during field operation and was subsequently measured under controlled dynamometer conditions. Thus, the recordings contain responses from an already damaged gearbox; they do not constitute a longitudinal record of fault initiation and progression.

The test turbine is a three-bladed, stall-controlled, upwind machine rated at 750 kW. The selected operating condition has a main-shaft speed of 22.09 rpm, nominal high-speed-shaft speed of 1,800 rpm, and electrical power equal to 50% of rated power. Vibration is sampled at 40 kHz. Eight accelerometer channels, AN3–AN10, are used; their locations are listed in Table 1. The benchmark associates the recordings with healthy or damaged gearbox condition. These condition labels provide the primary supervision for the binary detection task.

| Channel | Location |
|---|---|
| AN3 | Ring gear radial, 6 o'clock |
| AN4 | Ring gear radial, 12 o'clock |
| AN5 | LS-SH radial |
| AN6 | IMS-SH radial |
| AN7 | HS-SH radial |
| AN8 | HS-SH upwind bearing radial |
| AN9 | HS-SH downwind bearing radial |
| AN10 | Carrier downwind radial |

Table 1. Accelerometer channels used in the experiment. LS-SH, IMS-SH, and HS-SH denote the low-, intermediate-, and high-speed shafts, respectively.

Each 0.1 s segment contains 4,000 samples per channel. The study reports 50,000 segments, but this count requires reconciliation with the stated recording duration and segmentation rule. A 10-minute recording produces 6,000 non-overlapping 0.1 s windows per condition before channel counting. The number of synchronized graph examples therefore depends on the recordings selected, overlap, and exclusion criteria. This discrepancy does not change the tabulated results, but it prevents an exact reconstruction of the sample population from the current description.

## 4.2 Model configuration and training

The experiments compare GNN and GG-NN single-output models using sparse-filtered or manually engineered time- and frequency-domain features. SF-SoftMax serves as a non-graph baseline. The sequence experiments compare sparse filtering, raw segments, and manually engineered features within the GGS-NN setting.

The reported graph-model settings are hidden-state size 16, learning rate 0.001, dropout 0.2, and an L2 regularization coefficient of 0.05. Optimization uses Adam [22], with Glorot initialization [23] and a maximum of 1,000 epochs. The activation is described as log-sigmoid and the training loss as cross-entropy. Their exact placement, together with the output normalization, must be defined in the implementation because the hidden activation and the class-probability readout serve different roles.

The intended data partition is 60% training, 20% development, and 20% testing. An early-stopping interval of 20 epochs without loss improvement is reported, but the monitored loss and checkpoint selection need confirmation. Development loss should guide stopping and hyperparameter selection, with the test partition used only for final scoring. The propagation depth, batch size, random seeds, sparse-filtering optimizer, feature definitions, and computing hardware are necessary for replication and are not established by the current record.

### 4.3 Evaluation measures and scope

For binary health detection, accuracy is the proportion of correctly classified examples:

$$Accuracy = \frac{TP + TN}{TP + TN + FP + FN} \quad (7)$$

Here TP and TN denote correctly classified damaged and healthy examples, respectively, while FP and FN denote the corresponding errors. Class-specific recall, precision, F1 score, and confusion-matrix counts would provide a more complete assessment, particularly if class frequencies differ. Those quantities cannot be reconstructed from accuracy alone.

Sequence evaluation requires an additional definition. Token accuracy measures correctly predicted sequence elements, while exact-sequence accuracy requires all elements of an output to be correct. Component, relation, and final-state accuracy answer still different questions. The available sequence results are retained as reported accuracies because their scoring rule is not specified. They should not be directly compared with single-output classification accuracy as though both metrics measured the same event.

Data dependence also affects interpretation. Adjacent windows from a recording can be strongly related, and synchronized channels belong to the same physical observation. A recording-level or blocked temporal partition is preferable to distributing related windows across all partitions. Moreover, because the benchmark uses different physical gearboxes for the two conditions, the experiment does not separate all effects of gearbox identity from condition. Validation on additional gearboxes is needed to establish equipment-level generalization.

## 5 Results and discussion

### 5.1 Feature dimension

Table 2 compares four sparse-filtering representation sizes. Increasing K from 50 to 100 raises GNN accuracy from 74.24% to 92.33%, GG-NN accuracy from 88.25% to 93.74%, and GGS-NN accuracy from 82.57% to 90.73%. These changes indicate that 50 features provide a weaker representation for the evaluated models. They do not establish which physical signal properties are lost at that dimension.

| Features K | GNN (%) | GG-NN (%) | GGS-NN (%) | Time (s) |
|---|---|---|---|---|
| 50 | 74.24 | 88.25 | 82.57 | < 60 |
| 100 | 92.33 | 93.74 | 90.73 | < 60 |
| 300 | 91.72 | 94.14 | 90.12 | 62 |
| 500 | 92.41 | 93.62 | 90.83 | 78 |

Table 2. Reported accuracy for different sparse-filtering feature dimensions. One running-time value is supplied for each dimension; its attribution to a particular model or processing stage is unspecified.

Beyond K = 100, the accuracy changes are comparatively small and are not monotonic. The highest GG-NN accuracy, 94.14%, occurs at K = 300, whereas the highest GNN and GGS-NN accuracies occur at K = 500. Relative to K = 100, these best values differ by only 0.40, 0.08, and 0.10 percentage points, respectively. The reported times increase to 62 s at K = 300 and 78 s at K = 500. A 100-feature representation is therefore a reasonable compact setting for the reported comparison. Repeated trials are needed before treating the small accuracy differences as statistically meaningful.

The results do not demonstrate that higher dimensions cause overfitting. That explanation would require training and development curves or repeated measurements of the generalization gap. Similarly, the claim that a particular dimension isolates one active fault feature cannot be established from Table 2 alone. The defensible interpretation is that accuracy largely plateaus above 100 features in this experiment while the reported running time increases.

### 5.2 Single output classification

Table 3 presents the single-output results. SF-GG-NN achieves the highest reported accuracy, 93.74%, followed by SF-GNN at 92.33%. Relative to their manually engineered feature counterparts, sparse filtering improves GNN accuracy by 8.79 percentage points and GG-NN accuracy by 4.46 percentage points. The corresponding reported times fall from 58 to 54 s and from 52 to 47 s. These paired comparisons support the usefulness of the learned representation within the tested graph architectures.

| Model | Accuracy (%) | Time (s) |
|---|---|---|
| SF-GNN | 92.33 | 54 |
| SF-GG-NN | 93.74 | 47 |
| Feature-GNN | 83.54 | 58 |
| Feature-GG-NN | 89.28 | 52 |
| SF-SoftMax | 89.92 | 38 |

Table 3. Reported single-output classification accuracy and running time. Timing scope and hardware are not specified, so these values support only a descriptive comparison within the reported experiment.

SF-SoftMax achieves 89.92% accuracy with a reported time of 38 s. SF-GG-NN exceeds this accuracy by 3.82 percentage points and takes 9 s longer; SF-GNN exceeds it by 2.41 percentage points and takes 16 s longer. The graph models therefore offer a measured accuracy–time tradeoff relative to this baseline. The results do not support a general claim that graph processing is computationally cheaper than a non-graph classifier.

The differences are consistent with a benefit from combining learned features and graph processing, but the present comparison does not isolate the effect of equipment knowledge. Classifier architecture, parameter count, and graph structure all differ between a graph model and SoftMax regression. A matched non-graph model using the same sensor features, together with an edge-removal or edge-shuffling ablation, would clarify which part of the improvement comes from relational information.

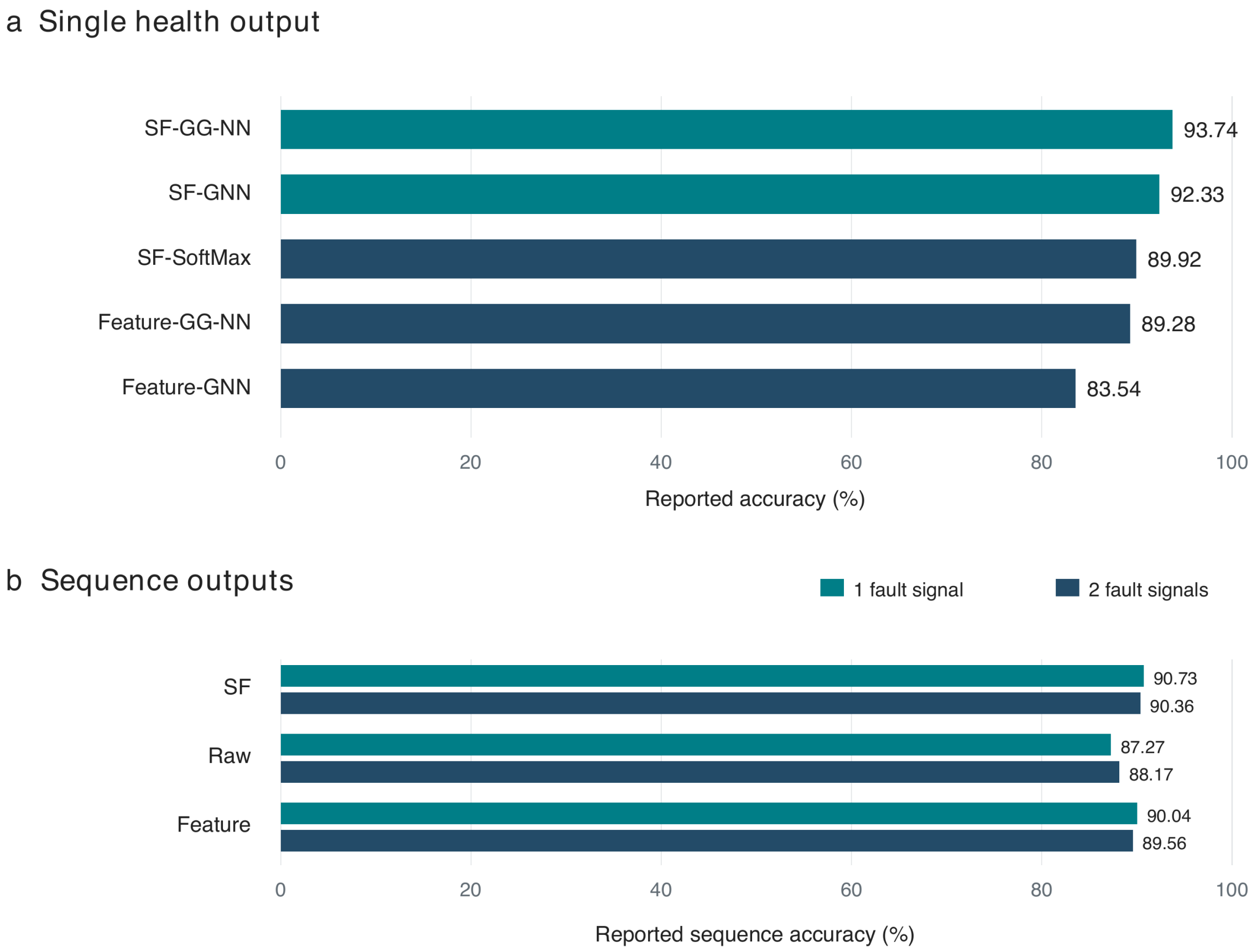

Figure 3. Visualization of the reported classification results. Panel a shows single-output accuracy from Table 3. Panel b shows sequence-output accuracy from Table 4. The panels retain separate output definitions and must not be interpreted as a direct comparison of identical metrics. No error bars are shown because repeated-run variability is unavailable.

### 5.3 Sequential outputs

Table 4 compares preprocessing methods for the one- and two-fault-signal cases. Sparse filtering gives reported accuracies of 90.73% and 90.36%, respectively. Compared with raw inputs, the gains are 3.46 percentage points for one fault signal and 2.19 percentage points for two fault signals. Compared with manually engineered features, the gains are 0.69 and 0.80 percentage points. The smaller margins over engineered features suggest that much of the benefit in the sequence setting is already captured by a suitable input representation.

| Preprocessing | 1 fault signal (%) | 2 fault signals (%) |
|---|---|---|
| SF | 90.73 | 90.36 |
| Raw | 87.27 | 88.17 |
| Feature | 90.04 | 89.56 |

Table 4. Reported sequence-output accuracy. The one- and two-fault-signal labels are retained from the experiment; the signal-selection procedure and sequence-scoring rule require specification.

For sparse filtering, the two-signal case is 0.37 percentage points below the one-signal case. This difference is small in the reported values, but the absence of repeated runs or uncertainty estimates prevents a conclusion of equivalent performance. Furthermore, two fault signals do not necessarily represent two independently introduced physical defects. The benchmark damaged gearbox contains multiple damage mechanisms, and the sequence experiment must document how each case was assembled and annotated.

Sequence outputs provide a structured link between a health prediction and the graph entities used to express it. Their usefulness should be assessed separately from the binary decision. A future evaluation should report errors by component, sensor, relation, and final health state, as well as exact-sequence accuracy. It should also distinguish relations already specified in the input graph from relations inferred by the decoder. This would show whether the sequence model learns additional diagnostic information or mainly reproduces known equipment structure.

### 5.4 Interpretation and limitations

The proposed framework offers two practical capabilities. Sparse filtering compresses each high-frequency segment before graph processing, and the graph retains the association between sensor features and equipment entities. Together, these capabilities provide a structured representation of multichannel observations. The reported results are strongest as evidence for classification within the selected benchmark. They do not establish early warning time, remaining useful life prediction, reduced maintenance cost, or improved grid reliability, because those outcomes were not evaluated.

The computational findings also require qualification. A lower-dimensional input can reduce the cost of subsequent learning, but sparse-filtering training and transformation introduce their own costs. The reported running times do not separate feature fitting, feature transformation, graph training, and inference. Without that separation and a hardware description, the experiment cannot substantiate real-time deployment or an end-to-end acceleration claim. Per-window inference latency and memory use would be more relevant deployment measures than a single aggregate running time.

The main validity concerns are the construction of the data partition, the unresolved sample count, possible dependence among adjacent windows, and the limited number of physical gearboxes. Uncertainty estimates are also unavailable. Consequently, the percentage-point differences describe the reported experiment and should not be generalized to a population of turbines. A stronger assessment would use repeated training runs, blocked or recording-level test partitions, class-specific metrics, and independent gearbox or operating-condition tests.

Recent graph-learning studies provide useful directions for extending this assessment [13–20]. Attention mechanisms and learned adjacency can test whether a fixed knowledge graph misses useful dependencies. Temporal graph models can distinguish within-window vibration features from changes across successive windows. Contrastive pretraining can address limited labels, while domain adaptation and uncertainty estimation can help assess unfamiliar operating conditions. These are extensions to be tested, rather than capabilities demonstrated by the present SF-GNN results.

## 6 Conclusions

We investigated a framework that combines sparse filtering with graph neural networks for vibration-based wind turbine gearbox fault detection. Sparse filtering supplies compact sensor features, and a knowledge-based graph connects those features to equipment and measurement entities. Basic and gated graph models provide a single health prediction, while a gated graph sequence model produces structured outputs associated with the graph.

At a feature dimension of 100, the reported SF-GG-NN accuracy is 93.74%, compared with 89.28% for its manually engineered feature counterpart and 89.92% for SF-SoftMax. Sparse filtering also improves the reported sequence accuracy over raw inputs, reaching 90.73% and 90.36% in the one- and two-fault-signal cases. These observations support further study of compact vibration features within a relational equipment representation.

The conclusions remain limited to the available benchmark and evaluation record. Resolving the segmentation count, documenting the full graph and decoding procedure, and verifying the independence of training and test data are necessary for reproducibility. Future experiments should isolate the contribution of graph relationships, quantify uncertainty and inference cost, and validate performance across additional gearboxes, speeds, and loads. Independent component-level annotations are required before interpreting semantic sequences as validated fault localization.

## References

[1] B. Samanta, "Gear fault detection using artificial neural networks and support vector machines with genetic algorithms," Mechanical Systems and Signal Processing, vol. 18, no. 3, pp. 625–644, 2004. https://doi.org/10.1016/S0888-3270(03)00020-7

[2] M. Amar, I. Gondal, and C. Wilson, "Vibration Spectrum Imaging: A Novel Bearing Fault Classification Approach," IEEE Transactions on Industrial Electronics, vol. 62, no. 1, pp. 494–502, 2015. https://doi.org/10.1109/TIE.2014.2327555

[3] R. K. Pandit and D. Infield, "SCADA-based wind turbine anomaly detection using Gaussian process models for wind turbine condition monitoring purposes," IET Renewable Power Generation, vol. 12, pp. 1249–1255, 2018. https://doi.org/10.1049/iet-rpg.2018.0156

[4] P. Bangalore and L. B. Tjernberg, "An Artificial Neural Network Approach for Early Fault Detection of Gearbox Bearings," IEEE Transactions on Smart Grid, vol. 6, no. 2, pp. 980–987, 2015. https://doi.org/10.1109/TSG.2014.2386305

[5] D. Zhang, L. Qian, B. Mao, C. Huang, B. Huang, and Y. Si, "A Data-Driven Design for Fault Detection of Wind Turbines Using Random Forests and XGboost," IEEE Access, vol. 6, pp. 21020–21031, 2018. https://doi.org/10.1109/ACCESS.2018.2818678

[6] X. Wu, G. Jiang, X. Wang, P. Xie, and X. Li, "A Multi-Level-Denoising Autoencoder Approach for Wind Turbine Fault Detection," IEEE Access, vol. 7, pp. 59376–59387, 2019. https://doi.org/10.1109/ACCESS.2019.2914731

[7] J. Wang, R. X. Gao, and R. Yan, "Integration of EEMD and ICA for wind turbine gearbox diagnosis," Wind Energy, vol. 17, no. 5, pp. 757–773, 2014. https://doi.org/10.1002/we.1653

[8] J. Ngiam, P. W. Koh, Z. Chen, S. Bhaskar, and A. Y. Ng, "Sparse Filtering," Advances in Neural Information Processing Systems 24, 2011. https://papers.neurips.cc/paper_files/paper/2011/file/192fc044e74dffea144f9ac5dc9f3395-Paper.pdf

[9] Y. Lei, F. Jia, J. Lin, S. Xing, and S. X. Ding, "An Intelligent Fault Diagnosis Method Using Unsupervised Feature Learning Towards Mechanical Big Data," IEEE Transactions on Industrial Electronics, vol. 63, no. 5, pp. 3137–3147, 2016. https://doi.org/10.1109/TIE.2016.2519325

[10] F. Scarselli, M. Gori, A. C. Tsoi, M. Hagenbuchner, and G. Monfardini, "The Graph Neural Network Model," IEEE Transactions on Neural Networks, vol. 20, no. 1, pp. 61–80, 2009. https://doi.org/10.1109/TNN.2008.2005605

[11] Y. Li, D. Tarlow, M. Brockschmidt, and R. Zemel, "Gated Graph Sequence Neural Networks," International Conference on Learning Representations, 2016. https://arxiv.org/abs/1511.05493

[12] D. Küçük and D. Küçük, “OntoWind: An Improved and Extended Wind Energy Ontology,” arXiv:1803.02808, 2018. https://arxiv.org/abs/1803.02808

[13] Q. Tan, P. Ma, H. Zhang, and N. Wang, “Fault Diagnosis for Wind Turbine Gearbox Based on Graph Attention Networks,” Acta Energiae Solaris Sinica, vol. 45, no. 1, pp. 265–274, 2024. https://doi.org/10.19912/j.0254-0096.tynxb.2022-1501

[14] J. Guo, C. Liu, S. Liu, and W. Liu, “Graph contrastive learning for semi-supervised wind turbine fault diagnosis with few labeled SCADA data,” Measurement, vol. 245, article 116531, 2025. https://doi.org/10.1016/j.measurement.2024.116531

[15] J. Ma, Y. Fu, T. Cheng, et al., “SCADA Data-Driven Spatio-Temporal Graph Convolutional Neural Network for Wind Turbine Fault Diagnosis,” IEEE Transactions on Instrumentation and Measurement, vol. 74, article 3521910, 2025. https://doi.org/10.1109/TIM.2025.3551875

[16] H. S. Kim and J. Lee, “Maintenance knowledge graph and time-series relational graph convolution neural networks for enhancing predictive maintenance of wind turbine gearbox,” Knowledge-Based Systems, article 115523, 2026. https://doi.org/10.1016/j.knosys.2026.115523

[17] D. Coronel, C. Guevara, and M. Santos, “Hybrid graph neural network with FFT and genetic optimization for fault detection in wind turbines,” Neurocomputing, vol. 684, article 133352, 2026. https://doi.org/10.1016/j.neucom.2026.133352

[18] Y. Fu, Y. Wu, D. He, Z. Jin, and B. Yu, “Spatio-temporal connected graph convolutional network for SCADA-data-based wind turbine fault diagnosis under time-varying operating conditions,” Journal of Vibration and Control, published online October 19, 2026. https://doi.org/10.1177/10775463261499353

[19] W. Zhou, Q. Zhou, and J. Zhang, “Speed–Load Insensitive Fault Diagnosis Method of Wind Turbine Gearbox Based on Adversarial Training,” Electronics, vol. 14, no. 4, article 732, 2025. https://doi.org/10.3390/electronics14040732

[20] J. Zhou and Y. Zhao, “A distance-aware approach for reliable out-of-distribution detection of wind turbine gearbox fault diagnosis,” Frontiers in Energy Research, vol. 12, article 1496130, 2024. https://doi.org/10.3389/fenrg.2024.1496130

[21] S. Sheng, “Wind Turbine Gearbox Vibration Condition Monitoring Benchmarking Datasets,” National Renewable Energy Laboratory, benchmark documentation, undated. https://data.openei.org/files/738/Vibration%20Condition%20Monitoring%20Benchmarking%20Datasets.pdf

[22] D. P. Kingma and J. Ba, “Adam: A Method for Stochastic Optimization,” International Conference on Learning Representations, 2015. https://arxiv.org/abs/1412.6980

[23] X. Glorot and Y. Bengio, “Understanding the difficulty of training deep feedforward neural networks,” Proceedings of the Thirteenth International Conference on Artificial Intelligence and Statistics, pp. 249–256, 2010. https://proceedings.mlr.press/v9/glorot10a.html